\documentclass{midl} 

\usepackage{booktabs, tabulary, multirow, multicol, makecell}
\usepackage{microtype}
\usepackage[utf8]{inputenc} 
\usepackage[table]{xcolor}
\usepackage[font=small]{caption}
\definecolor{Gray}{gray}{0.9}
\usepackage{mwe} 
\jmlrvolume{-- Under Review}
\jmlryear{2026}
\jmlrworkshop{Full Paper -- MIDL 2026 submission}
\editors{Under Review for MIDL 2026}

\title[Generalizing Abstention]{Generalizing Abstention for Noise-Robust Learning in Medical Image Segmentation}

\midlauthor{\Name{Wesam Moustafa\nametag{$^{1,2}$}}\Email{wesam.h.moustafa@gmail.com}\\
\Name{Hossam Elsafty\nametag{$^{1,2,3}$}}\orcid{0009-0001-4095-4912}\Email{hossam.elsafty@iais.fraunhofer.de}\\
\Name{Helen Schneider\nametag{$^{1,2,3}$}}\Email{helen.schneider@iais.fraunhofer.de}\\
\Name{Lorenz Sparrenberg\nametag{$^{1,3}$}}\orcid{0000-0001-9450-7387}\Email{sparrenberg@bit.uni-bonn.de}\\
\Name{Rafet Sifa\nametag{$^{1,2,3}$}}\Email{rafet.sifa@iais.fraunhofer.de}\\
\addr $^{1}$ Institute of Computer Science, University of Bonn, Bonn, Germany \\
\addr $^{2}$ Fraunhofer Institute for Intelligent Analysis and Information Systems IAIS, Sankt Augustin, Germany \\
\addr $^{3}$ Lamarr Institute for Machine Learning and Artificial Intelligence, Dortmund, Germany
}

\begin{document}

\maketitle

\begin{abstract}
Label noise is a critical problem in medical image segmentation, often arising from the inherent difficulty of manual annotation. Models trained on noisy data are prone to overfitting, which degrades their generalization performance. While a number of methods and strategies have been proposed to mitigate noisy labels in the segmentation domain, this area remains largely under-explored. The abstention mechanism has proven effective in classification tasks by enhancing the capabilities of Cross Entropy, yet its potential in segmentation remains unverified. In this paper, we address this gap by introducing a universal and modular abstention framework capable of enhancing the noise-robustness of a diverse range of loss functions. 
Our framework improves upon prior work with two key components: an informed regularization term to guide abstention behaviour, and a more flexible power-law-based auto-tuning algorithm for the abstention penalty. We demonstrate the framework's versatility by systematically integrating it with three distinct loss functions to create three novel, noise-robust variants: GAC, SAC, and ADS. Experiments on the CaDIS and DSAD medical datasets show our methods consistently and significantly outperform their non-abstaining baselines, especially under high noise levels. This work establishes that enabling models to selectively ignore corrupted samples is a powerful and generalizable strategy for building more reliable segmentation models. Our code is publicly available at \url{https://github.com/wemous/abstention-for-segmentation}.
\end{abstract}


\section{Introduction}

The remarkable advancements in deep learning have revolutionized numerous fields, largely propelled by the availability of labelled datasets \cite{karimiDeepLearningNoisy2020, zhangDisentanglingHumanError2020, garcia-garciaReviewDeepLearning2017}. However, the presence of label noise is a significant impediment to the generalizability of Deep Neural Networks (DNNs), as their immense capacity makes them prone to memorizing incorrect labels, which harms their ability to generalize to unseen data \cite{lienen2024mitigating, gonzalez-santoyoIdentifyingMitigatingLabel2025, schneider2024potential, schneider2023one}. This problem is especially pronounced in medical image segmentation, where obtaining clean, pixel-level annotations is notoriously difficult and expensive, and where annotation errors can have direct clinical consequences \cite{karimiDeepLearningNoisy2020, zhangDisentanglingHumanError2020, xuAdvancesMedicalImage2024}. Training on such noisy labels leads to incorrect gradients, causing the model to learn erroneous patterns and fail in critical applications \cite{marcinkiewiczQuantitativeImpactLabel2019}.

To counteract label noise, a variety of robust learning methodologies have been developed, primarily for classification tasks. These include noise filtering techniques \cite{gonzalez-santoyoIdentifyingMitigatingLabel2025}, loss reweighting strategies \cite{karimiDeepLearningNoisy2020}, and curriculum learning \cite{lienen2024mitigating}. While promising, these methods often introduce computational complexity or require strong assumptions about the noise characteristics \cite{gonzalez-santoyoIdentifyingMitigatingLabel2025}. Among the explored directions, noise-robust loss functions are a compelling alternative due to their simplicity, efficiency, and model-agnostic nature \cite{staatsEnhancingNoiseRobustLosses2025}. By leveraging properties like boundedness \cite{gce} and symmetry \cite{sce}, they modify the optimization objective to inherently limit the influence of noisy examples and prevent overfitting \cite{toner2023label, ding2024improve}.

Despite these advances, a research gap remains for robust learning specifically in image segmentation, where existing methods often struggle to address the spatially-correlated inaccuracies inherent in annotation noise \cite{guoImbalancedMedicalImage2025, karimiDeepLearningNoisy2020}. In this paper, we propose to address this gap by adapting and generalizing the \textbf{abstention} mechanism, a powerful technique that has proven effective in mitigating label noise in classification \cite{karimiDeepLearningNoisy2020, dac, idac}. The abstention mechanism empowers a DNN to abstain from making a prediction on confusing or unreliable samples by integrating an abstention option directly into the training process. Building upon the foundational Deep Abstaining Classifier (DAC) \cite{dac} and its extension, the Informed Deep Abstaining Classifier (IDAC) \cite{idac}, this paper makes several contributions to advance noise-robust medical image segmentation:

\begin{itemize}
    \item \textbf{Adaptation of Abstention to Segmentation}: We investigate the applicability of the abstention mechanism to image segmentation by adapting the DAC and IDAC loss functions for this domain.
    \item \textbf{Enhanced and Generalized Abstention Definition}: Our contribution improves and generalizes abstention, incorporating an informed regularization term guided by estimated noise rates \(\tilde\eta\) and a power-law-based \(\alpha\) auto-tuning algorithm.   
    \item \textbf{Loss-Agnostic Integration and Novel Loss Functions}: We integrate the enhanced abstention mechanism with other loss functions, including Generalized Cross Entropy (GCE) \cite{gce}, Symmetric Cross Entropy (SCE) \cite{sce}, and Dice Loss \cite{dice}, introducing three loss functions: the Generalized Abstaining Classifier (GAC), the Symmetric Abstaining Classifier (SAC), and the Abstaining Dice Segmenter (ADS). ADS introduces architectural adaptations for class-wise abstention and class-specific noise rates \(\tilde\eta_c\).  
    \item \textbf{Empirical Validation of Robustness and Versatility}: Through empirical evaluations and quantitative analysis (\figureref{fig:intro}) on medical image datasets (CaDIS \cite{cadis} and DSAD \cite{dsad}) under varying noise levels, we show consistent superiority over non-abstaining baselines.  
\end{itemize}
The remainder of this paper reviews related work (\sectionref{sec:related}), details our proposed abstention framework and novel loss functions (\sectionref{sec:method}), outlines the experimental setup (\sectionref{sec:exp}), and presents a comprehensive evaluation of our results (\sectionref{sec:results}) before concluding in \sectionref{sec:conclusions}.

\begin{figure}[htbp]
    \figureconts
    {fig:intro}
    {\caption{The impact of our noise-robust abstention framework. On a CaDIS sample with 25\% label noise, the baseline Dice Loss (b) produces a noisy and inaccurate mask. In contrast, our proposed \textbf{Abstaining Dice Segmenter (ADS)} (c) yields a result that is visually cleaner and adheres more closely to the ground truth (a).}}
    {
    \subfigure[Ground Truth]{
    \label{fig:intro-gt}
    \includegraphics[width=0.25\textwidth]{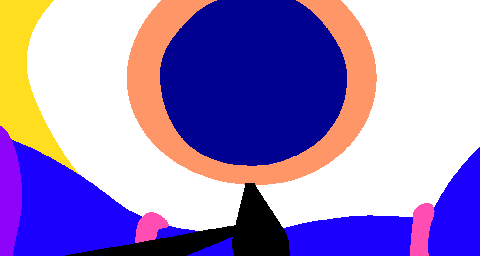}
    }
    \subfigure[Dice]{
    \label{fig:intro-dice}
    \includegraphics[width=0.25\textwidth]{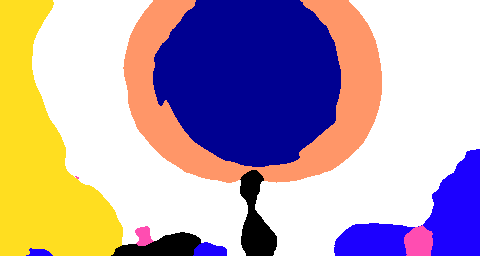}
    }
    \subfigure[\bfseries ADS]{
    \label{fig:intro-dads}
    \includegraphics[width=0.25\textwidth]{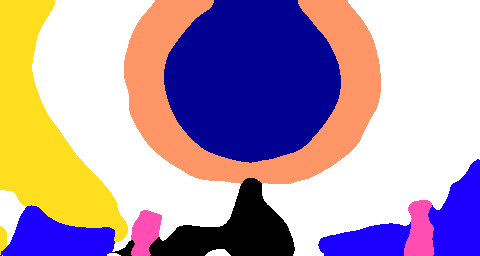}
    }
    }
\end{figure}


\section{Related Work} \label{sec:related}
The deployment of deep learning in medical imaging is frequently hampered by data imperfections, ranging from data scarcity and class imbalance \cite{tomar2025first, tomar2025effective} to missing or uncertain annotations \cite{schneider2023spml}. Within this landscape, label noise remains a particularly pervasive challenge that has been explored through several avenues \cite{karimiDeepLearningNoisy2020}. 
Early strategies operated at different granularities, seeking to identify and correct noisy labels. These included pixel-wise adaptive weight maps, as proposed by \citet{shiDistillingEffectiveSupervision2021}, which dynamically adjust the contribution of each pixel to the loss, and graph-based label correction by \citet{yiLearningPixelLevelLabel2022}, which models spatial relationships to propagate corrections from reliable to unreliable pixels. Other frameworks operated at the image level to assess overall annotation quality, with some combining both pixel- and image-level perspectives to distill supervision more effectively \cite{shiDistillingEffectiveSupervision2021,zhuPickandLearnAutomaticQuality2019}. Recognizing that segmentation noise is often not random but spatially correlated, other works have proposed explicit noise models. The Markov models from \citet{yaoLearningSegmentNoisy2023}, for instance, simulate realistic boundary distortions, while methods like LVC-Net by \citet{shuLVCNetMedicalImage2019} leverage local visual cues to guide the network away from incorrect labels during training.

A significant approach exploits the intrinsic learning dynamics of deep networks, particularly the `early-learning' phenomenon, where models tend to fit clean, simple patterns before eventually memorizing the noise present in incorrect labels \cite{liuAdaptiveEarlyLearningCorrection2022,yeActiveNegativeLoss2024}. This observation has led to the development of adaptive correction methods like ADELE by \citet{liuAdaptiveEarlyLearningCorrection2022}, which detects the onset of memorization for each semantic class to intervene at the optimal moment. In a similar vein, multi-network and co-training paradigms leverage the consensus or disagreement between two or more models to filter out noisy signals. By using diverse architectures, these methods reduce the risk of confirmation bias, where a single model reinforces its own errors \cite{liSemiSupervisedSemanticSegmentation2023,rongBoundaryenhancedCotrainingWeakly2023}.

Furthermore, emerging paradigms reframe the problem by treating large-scale noisy labels not as a hindrance but as a valuable resource. Pretraining strategies use massive, imperfectly labelled datasets to learn robust feature representations that can be fine-tuned on smaller, clean datasets \cite{liuCromSSCrossmodalPretraining2025}. Other methods use meta-learning to bootstrap robust models, such as L2B by \citet{zhouL2BLearningBootstrap2024}, which learns to dynamically weight the influence of observed labels and model-generated pseudo-labels during training.


An alternative and more fundamental approach that is most relevant to our work involves designing inherently robust loss functions. Instead of relying on external modules for noise detection or correction, this strategy embeds noise tolerance directly into the optimization objective \cite{karimiDeepLearningNoisy2020}. Examples include the T-Loss from \citet{gonzalez-jimenezRobustTLossMedical2023}, which is based on the heavy-tailed Student-t distribution to reduce the influence of outliers, and the Active Negative Loss (ANL) framework proposed by \citet{yeActiveNegativeLoss2024}. Our work contributes to this line of research, but with a distinct and more modular philosophy. Instead of designing a new loss function from scratch, we propose a modular mechanism that can enhance the inherent robustness of existing losses. \\

A promising strategy for mitigating the impact of label noise is to empower a model to abstain from making a prediction on samples it deems unreliable. This approach circumvents the core problem of standard supervised learning, where a model is forced to commit to a prediction, potentially leading it to memorize erroneous labels. The concept was formally introduced for deep learning in the Deep Abstaining Classifier (DAC) by \citet{dac}. The DAC framework enables abstention by augmenting a network's architecture with an additional \((k+1)\)-th output neuron, which explicitly represents the choice to abstain. Its corresponding loss function is defined as:
\begin{equation}\label{eq:dac}
    \mathcal{L}_{DAC}(x_j)  = (1-p_{k+1})\left(-\sum^k_{i=1} t_i \log\frac{p_i}{1-p_{k+1}}\right) + \alpha\log\frac{1}{1-p_{k+1}}
\end{equation}

The loss is composed of two competing terms. The first is a modified Cross Entropy (CE) loss scaled by \((1-p_{k+1})\), which represents the confidence in \textit{not} abstaining, while the classification probability \(p_i\) for each class \(i\) is re-normalized by the same factor. 
The second component is a regularization term that directly penalizes the act of abstaining, where \(p_{k+1}\) is the model's predicted probability of abstention \cite{dac}.

The abstention penalty is controlled by the hyperparameter \(\alpha\). Crucially, DAC employs an adaptive auto-tuning schedule where, after an initial warm-up period, \(\alpha\) is initialized to a small value and is linearly increased over the remaining epochs up to a predefined value, \(\alpha_{final}\). This ramp-up strategy, detailed in \appendixref{app:alpha}, acts as a form of curriculum learning, initially permitting the model to ignore noisy samples and progressively forcing it to learn from more challenging data as its confidence grows \cite{dac}.



Building directly upon this foundation, \citet{idac} proposed the Informed Deep Abstaining Classifier (IDAC) to create a more targeted response to noise. IDAC refines the abstention mechanism by incorporating an a priori estimation of the dataset's noise level, \(\tilde\eta\), directly into its regularization term. The IDAC loss functions is defined as:
\begin{equation}\label{eq:idac}
    \mathcal{L}_{IDAC}(x_j)=(1-p_{k+1})\left(-\sum^k_{i=1} t_i\log\frac{p_i}{1-p_{k+1}}\right)+\alpha(\tilde\eta-\hat\eta)^2
\end{equation}

The key innovation lies in replacing DAC's incremental penalty with a term that minimizes the divergence between the expected noise rate \(\tilde\eta\) and the model's current batch-wise abstention rate, \(\hat\eta\). This provides a more direct supervisory signal, guiding the model to abstain on a fraction of samples that is consistent with the known level of label corruption \cite{idac}.

While DAC and IDAC have demonstrated the profound effectiveness of abstention, their application has been confined to the CE loss paradigm. Our work addresses this limitation by proposing a generalized abstention framework, establishing it as a modular tool to enhance the robustness of a diverse range of loss functions.

\section{The Universal Abstention Framework}\label{sec:method}
Building upon the demonstrated efficacy of DAC and IDAC in mitigating label noise through abstention, we propose an enhanced and universal definition of the abstention mechanism that can be readily adapted to virtually any underlying loss function, \(\mathcal{L}_X(x_j)\). Our generalized abstaining loss is formulated as:
\begin{equation}\label{eq:abstention}
    \mathcal{L}_{abstention}(x_j) = (1-p_{k+1})\mathcal{L}_{X}(x_j) + \alpha\left|\log\frac{1-\tilde\eta}{1-p_{k+1}}\right|
\end{equation}
With this formulation, we introduce two critical innovations designed to provide greater flexibility and more targeted noise mitigation.
\subsection{Informed Regularization}
The first improvement lies in the \textbf{regularization term} \(\alpha\left|\log\frac{1-\tilde\eta}{1-p_{k+1}}\right|\). This term draws inspiration from IDAC by explicitly incorporating the expected noise rate \(\tilde\eta\) to guide the abstention behaviour. Unlike DAC, which pushes the abstention probability \(p_{k+1}\) toward zero, our term incentivises the model to maintain \(p_{k+1}\) in proximity to \(\tilde\eta\). This allows the model to continue abstaining on samples it confidently perceives as noisy, rather than being forced to make classification decisions that could elevate the risk of overfitting to noise. This enhanced definition of the regularization term is also flexible; if a reliable estimate for \(\tilde\eta\) isn't available, setting \(\tilde\eta=0\) effectively reduces the term to its original DAC form, which has already demonstrated its strength and effectiveness in combating label noise.

\subsection{Power-Law Auto-Tuning}\label{sec:gamma}
The second and more significant enhancement concerns the \textbf{\(\boldsymbol{\alpha}\) auto-tuning algorithm}. The original algorithm proposed by DAC employed a linear ramp-up strategy for \(\alpha\) after a warm-up phase, which, while effective, offered limited flexibility in controlling the learning trajectory. Our refined approach replaces this with a simpler yet more powerful and flexible method. For every epoch \(e\) after an initial warm-up phase of \(L\) epochs out of a total \(E\) epochs, \(\alpha\) is dynamically calculated as:
\begin{equation}\label{eq:alpha}
    \alpha = \alpha_{final} * \left(\frac{e-L}{E-L}\right)^\gamma
\end{equation}
In this equation, \(\gamma>0\) serves as a growth factor that precisely controls the rate at which \(\alpha\) increases throughout the abstention phase, as depicted in \appendixref{app:gamma}. The behaviour of \(\alpha\) is modulated by \(\gamma\): if \(\gamma>1\) , \(\alpha\) exhibits a sublinear growth, increasing slowly at the beginning of the abstention period and accelerating its growth towards the end of training. This behaviour intensifies with larger values of \(\gamma\). Conversely, if \(\gamma<1\), \(\alpha\) experiences superlinear growth early in the abstention phase, with its rate of increase slowing down as training progresses. Setting \(\gamma=1\) yields a linear increment, akin to DAC's approach. This formulation provides significant flexibility in penalizing and guiding the abstention behaviour, enabling a more optimal balance between the model's learning from clean data and its strategic abstention from noisy or ambiguous samples. 

\subsection{Novel Abstaining Loss Functions for Segmentation}
We demonstrate our framework's versatility by creating three novel, noise-robust loss functions, one of which is tailored for segmentation.

\subsubsection{Abstaining Classifiers (GAC and SAC)} 
We first integrate our framework with two CE-based losses. The \textbf{G}eneralized \textbf{A}bstaining \textbf{C}lassifier (GAC) combines abstention with Generalized Cross Entropy (GCE) \cite{gce}, creating a dual defence where GCE's bounded loss attenuates noise on classified samples, while abstention filters out the most corrupted ones. The \textbf{S}ymmetric \textbf{A}bstaining \textbf{C}lassifier (SAC) enhances Symmetric Cross Entropy (SCE) \cite{sce}, empowering the model to completely disengage from highly suspect samples, rather than merely re-balancing their influence. SAC can actively filter out the most egregious noisy examples, allowing the symmetrical CE-RCE components to focus on refining predictions for the more reliable data. 

\subsubsection{Abstaining Dice Segmenter (ADS)}
\begin{figure}[htb]
    \floatconts
    {fig:ads_output}
    {\caption{Transforming the output layer from standard pixel-wise abstention (a) to our proposed class-wise abstention head for \textbf{ADS} (b). The dimensions \(b, c, h, w\) represent batch size, number of classes, height, and width, respectively. For the Adaptive Average Pool layer, \(w, w\) is the output size}}
    {
    \subfigure[Pixel-wise abstention]{
        \label{fig:output_a}
        \includegraphics[width=0.4\textwidth]{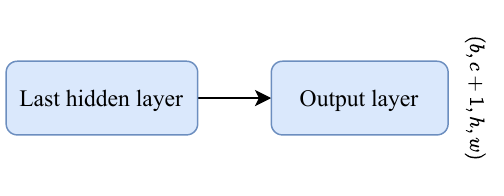}
    }
    \subfigure[Class-wise abstention]{
        \label{fig:output_c}
        \includegraphics[width=0.55\textwidth]{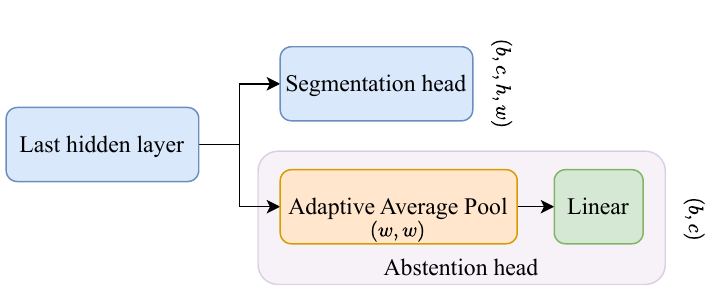}
    }
}
\end{figure}

Our most significant adaptation is the \textbf{A}bstaining \textbf{D}ice \textbf{S}egmenter (ADS)\footnote{The \textit{Segmenter} in ADS highlights its design for segmentation tasks, contrasting with the other 'Classifier' losses which can also be used for classification.}, which integrates our framework with the region-based Dice loss. This required two fundamental architectural changes to resolve the incompatibility between Dice's class-wise nature and standard pixel-wise abstention.:
\begin{itemize}
    \item \textbf{Class-wise Abstention Head:} We re-conceptualized the network's output to produce class-wise abstention predictions. As illustrated in \figureref{fig:output_c}, a specialized module uses Adaptive Average Pooling with an output size \(w*w\), followed by a Linear layer and \textit{sigmoid} activation to output a unique abstention probability for each of the \(k\) classes.
    \item \textbf{Class-specific Regularization:} To complement the class-wise abstention, the regularization term in \equationref{eq:abstention} is modified to use a vector of class-specific noise estimates, \(\boldsymbol{\tilde\eta_c}\), instead of a global noise estimation \(\tilde\eta\). This provides granular control over the abstention behaviour for each class, which is crucial for complex medical segmentation tasks.
\end{itemize}

\section{Experimental Setup}\label{sec:exp}
We validated our framework on two publicly available surgical datasets with distinct characteristics to demonstrate the generalizability of our approach.
\begin{itemize}
    \item Cataract Dataset for Image Segmentation (CaDIS): A benchmark featuring 4,670 frames from cataract surgery with dense, high-quality pixel-wise annotations \cite{cadis}. For our experiments, we utilized the 8-class variant, which groups surgical tools into a single class. Images were normalized and resized to 480x256.
    \item Dresden Surgical Anatomy Dataset (DSAD): A more complex benchmark with 1,430 frames from laparoscopic surgery \cite{dsad}. This dataset presents a greater challenge due to its intricate anatomical structures, sparse annotations (approx. 82\% background), and significant class imbalance. Images were normalized resized to 480x384.
\end{itemize}

To rigorously assess noise robustness, we simulated realistic annotation errors with a two-pronged approach. 
Structural noise was introduced via morphological transformations (erosion and dilation) to simulate boundary inaccuracies, while semantic noise was injected via stochastic label flipping to mimic annotator bias \cite{karimiDeepLearningNoisy2020, zhangDisentanglingHumanError2020, marcinkiewiczQuantitativeImpactLabel2019, liSemiSupervisedSemanticSegmentation2023}. 
We evaluated performance across five calibrated noise levels for each dataset: 5-25\% corruption for CaDIS and 3-15\% for DSAD.

We used an NVIDIA A100 80GB to train an U-Net model \cite{unet} with a pretrained ResNet-50 \cite{resnet} backbone for our experiments. Key training hyperparameters are detailed in \appendixref{app:train_params}. To ensure statistical reliability, each experiment was conducted five times with distinct random seeds. Hyperparameters for each loss function were optimized to yield the highest validation mean Intersection over Union (mIoU) on the highest noise level for each dataset, thereby maximizing noise resistance. Crucially, to ensure a fair comparison and rigorously isolate the impact of our abstention framework, the optimal hyperparameters found for the baseline GCE and SCE functions were deliberately held constant for their respective abstaining versions, GAC and SAC. While likely suboptimal for our novel functions, this methodology ensures that any observed performance gains are attributable solely to the abstention mechanism itself. \appendixref{app:loss_params} details the hyperparameters we used for each loss in our benchmarks.

\section{Evaluations and Visual Analysis}\label{sec:results}

The efficacy of our proposed framework was validated through extensive experiments on the CaDIS and DSAD datasets. The results, summarized in \tableref{tab:main} and \figureref{fig:results}, demonstrate the superior noise-robustness conferred by the abstention mechanism.
As anticipated, all loss functions exhibited a performance decline with increasing label noise, underscoring the universal challenge of learning from corrupted data. However, the critical distinction lies in the rate of this degradation. Our proposed abstaining loss functions consistently demonstrated a more graceful performance decline and maintained a significant advantage over their respective non-abstaining baselines, particularly at high noise intensities. On the CaDIS dataset at 25\% noise, our Abstaining Dice Segmenter (ADS) emerged as the top performer, achieving a 5.35\% mIoU lead over the standard Dice Loss. Similarly, GAC and SAC surpassed their baselines, confirming the broad applicability of our framework. This trend persisted on the more complex DSAD dataset, where despite lower overall mIoU scores, the abstaining variants maintained a clear and consistent performance advantage, highlighting their robust effectiveness even in challenging segmentation scenarios. The flatter degradation curves for the abstaining methods in \figureref{fig:results} highlight their superior resilience.
\begin{table}[htb]
    \renewcommand{\arraystretch}{1.25}
    \tableconts
    {tab:main}
    {\caption{Average test mIoU~(\%) and standard deviation across 5 runs of a U-Net model trained on CaDIS and DSAD datasets. \textbf{Gray background:} Abstaining loss functions. \textbf{(*):} Our proposed novel loss functions. \textbf{Structure:} The table is divided into four comparative groups (separated by double vertical lines); each group compares a baseline loss against its abstaining counterpart(s). \textbf{Bold:} Indicates the best result \textit{within that specific group}. For example, in the last group, we compare Dice vs. ADS to isolate the impact of our framework on the Dice loss.}}
    {
    \resizebox{\textwidth}{!}{
    \begin{tabulary}{\textwidth}{c|c|c>{\columncolor{Gray}}c>{\columncolor{Gray}}c||c>{\columncolor{Gray}}c||c>{\columncolor{Gray}}c||c>{\columncolor{Gray}}c}
    \hline
    \multirow{2}{*}{Dataset} & \multirow{2}{*}{\makecell{Noise rate \\ $\eta$~(\%)}} & \multicolumn{9}{c}{Loss function} \\
    & & CE & DAC & IDAC & GCE & GAC* & SCE & SAC* & Dice & ADS* \\
    \hline
    \multirow{6}{*}{CaDIS} & 0 & \textbf{76.02$\pm$0.70} & 75.29$\pm$0.79 & 75.36$\pm$0.73 & 73.49$\pm$3.27 & \textbf{73.76$\pm$2.80} & 75.38$\pm$0.75 & \textbf{75.83$\pm$0.62} & 76.52$\pm$0.47 & \textbf{77.04$\pm$0.37} \\
    & 5 & \textbf{73.67$\pm$1.03} & 73.14$\pm$0.46 & 72.89$\pm$0.41 & \textbf{72.83$\pm$1.11} & 71.73$\pm$2.79 & 73.41$\pm$0.71 & \textbf{73.51$\pm$1.59} & 73.48$\pm$0.28 & \textbf{75.22$\pm$0.85} \\
    & 10 & 66.39$\pm$0.17 & \textbf{67.43$\pm$0.49} & 66.92$\pm$0.49 & \textbf{64.82$\pm$0.86} & 64.16$\pm$2.57 & 65.92$\pm$0.91 & \textbf{67.29$\pm$1.65} & 66.51$\pm$0.61 & \textbf{71.12$\pm$0.55} \\
    & 15 & 64.15$\pm$2.47 & \textbf{65.85$\pm$1.05} & 64.87$\pm$0.91 & \textbf{64.81$\pm$0.46} & 64.44$\pm$2.70 & 62.16$\pm$1.99 & \textbf{65.48$\pm$2.11} & 67.31$\pm$0.73 & \textbf{70.80$\pm$1.08} \\
    & 20 & 59.56$\pm$1.21 & \textbf{63.42$\pm$0.87} & 60.54$\pm$2.27 & 60.73$\pm$1.41 & \textbf{60.91$\pm$1.64} & 57.62$\pm$4.22 & \textbf{62.70$\pm$0.31} & 63.64$\pm$0.82 & \textbf{68.88$\pm$0.49} \\
    & 25 & 52.27$\pm$1.70 & \textbf{60.63$\pm$2.73} & 58.19$\pm$4.77 & 55.71$\pm$1.30 & \textbf{59.46$\pm$0.76} & 55.08$\pm$0.93 & \textbf{61.27$\pm$1.22} & 61.04$\pm$1.41 & \textbf{66.39$\pm$0.67} \\
    \hline
    \hline
    \multirow{6}{*}{DSAD} & 0 & \textbf{34.25$\pm$2.50} & 34.01$\pm$0.96 & 33.60$\pm$0.72 & \textbf{35.14$\pm$1.65} & 32.26$\pm$0.53 & 32.78$\pm$1.19 & \textbf{33.86$\pm$1.83} & \textbf{31.28$\pm$0.87} & 30.09$\pm$1.10 \\
    & 3 & \textbf{33.69$\pm$1.85} & 33.67$\pm$2.01 & 32.76$\pm$2.03 & \textbf{33.84$\pm$2.56} & 32.94$\pm$2.23 & \textbf{32.11$\pm$1.09} & 30.90$\pm$2.76 & \textbf{30.83$\pm$4.78} & 28.64$\pm$2.76 \\
    & 6 & \textbf{30.70$\pm$2.47} & 29.47$\pm$1.97 & 29.11$\pm$2.10 & 29.69$\pm$1.96 & \textbf{29.78$\pm$4.27} & 30.51$\pm$2.16 & \textbf{31.55$\pm$2.43} & 28.56$\pm$1.00 & \textbf{30.48$\pm$3.61} \\
    & 9 & \textbf{24.65$\pm$2.90} & 24.58$\pm$2.61 & 23.47$\pm$2.48 & 22.95$\pm$2.93 & \textbf{28.84$\pm$4.17} & 28.02$\pm$2.37 & \textbf{28.55$\pm$1.29} & 19.04$\pm$1.92 & \textbf{26.23$\pm$2.05} \\
    & 12 & 21.00$\pm$3.15 & \textbf{22.59$\pm$4.35} & 20.94$\pm$1.86 & 19.84$\pm$2.89 & \textbf{25.00$\pm$4.13} & 21.57$\pm$0.67 & \textbf{23.73$\pm$0.68} & 16.15$\pm$1.49 & \textbf{22.63$\pm$0.51} \\
    & 15 & 14.41$\pm$2.59 & \textbf{17.69$\pm$3.97} & 16.24$\pm$1.45 & 14.12$\pm$2.91 & \textbf{20.01$\pm$2.56} & 15.31$\pm$0.75 & \textbf{15.91$\pm$3.53} & 14.65$\pm$1.50 & \textbf{18.05$\pm$1.63} \\
    \hline
    \end{tabulary}
    }
    }
\end{table}

A qualitative review of the segmentation masks, depicted in \figureref{fig:cadis-vis,fig:dsad-vis} further substantiates these quantitative gains. Visual inspection of the CaDIS results revealed that models trained with our abstaining losses produced markedly cleaner and more accurate segmentations than their baselines. Contours were sharper, noise artifacts were reduced, and overall structural coherence was improved. Most notably, ADS produced masks with the highest similarity the ground truth even at the highest noise level. While the inherent difficulty of the DSAD dataset resulted in lower-quality predictions across all methods, the same relative improvements were observed. The abstaining models consistently generated more coherent masks with fewer spurious predictions and better-defined boundaries compared to their non-abstaining counterparts. This visual evidence confirms that the improvements in mIoU translate directly to more reliable and clinically relevant segmentation outputs.

\begin{figure}[htb]
    \figureconts
    {fig:results}
    {\caption{Quantitative comparison of noise-robustness. The plots show the average test mIoU~(\%) degradation as label noise \(\eta\) increases. The flatter curves of the abstaining variants (solid lines) demonstrate their superior resilience compared to non-abstaining baselines (dashed lines). Our proposed losses (GAC, SAC, ADS) are in \textbf{bold}.}}
    {
    \subfigure[CaDIS Dataset]{
        \label{fig:results cadis}
        \includegraphics[width=0.47\textwidth]{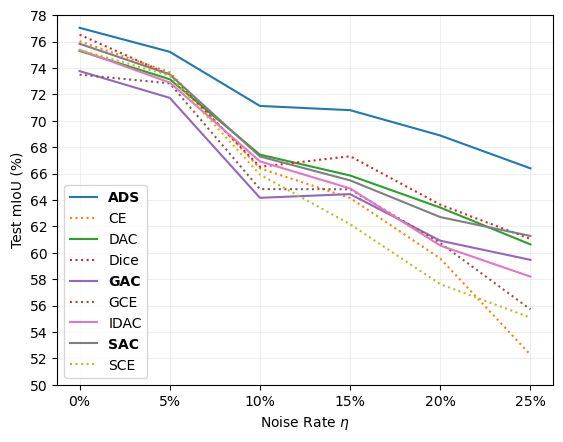}
    }
    \subfigure[DSAD Dataset]{
        \label{fig:results dsad}
        \includegraphics[width=0.47\textwidth]{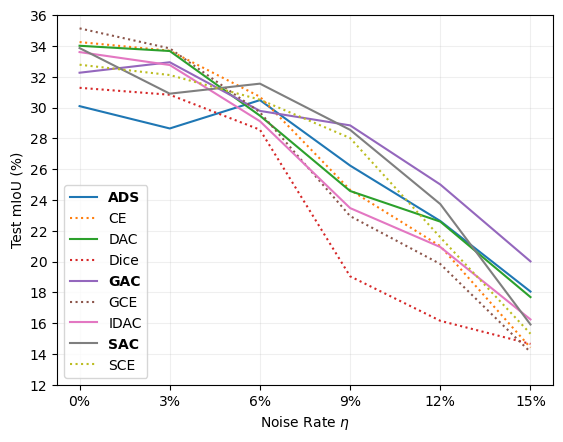}
    }
    }
\end{figure}

\begin{figure}[htb]
    \figureconts
    {fig:cadis-vis}
    {\caption{Qualitative comparison on a CaDIS sample at 25\% noise. Our proposed abstaining losses (\textbf{GAC}, \textbf{SAC}, \textbf{ADS}) produce masks with higher fidelity and fewer artifacts than their baselines. Abstaining losses are in \textbf{bold}.}}
    {
    \subfigure[Ground truth]{
        \label{fig:cadis-gt}
        \includegraphics[width=0.175\textwidth]{samples/cadis-gt.png}
    }
    \subfigure[CE]{
        \label{fig:cadis-ce}
        \includegraphics[width=0.175\textwidth]{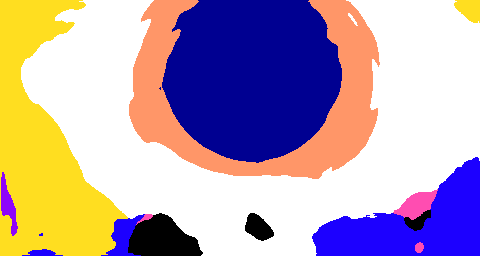}
    }
    \subfigure[\bfseries DAC]{
        \label{fig:cadis-dac}
        \includegraphics[width=0.175\textwidth]{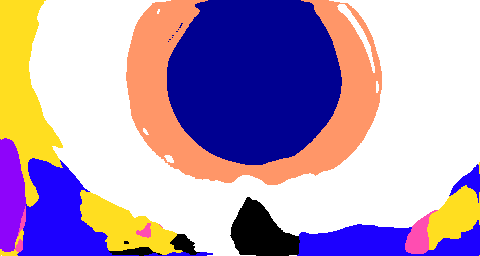}
    }
    \subfigure[\bfseries IDAC]{
        \label{fig:cadis-idac}
        \includegraphics[width=0.175\textwidth]{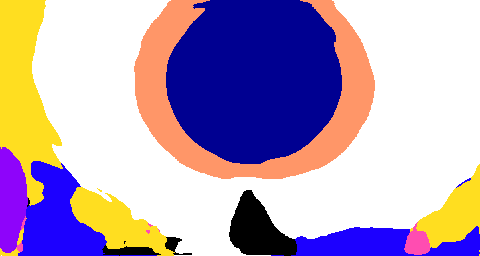}
    }
    \subfigure[GCE]{
        \label{fig:cadis-gce}
        \includegraphics[width=0.175\textwidth]{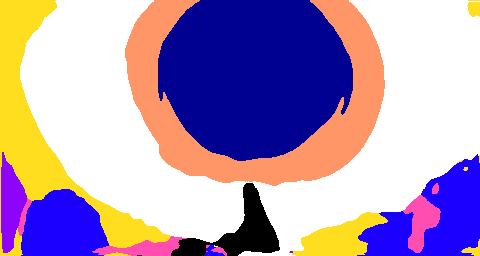}
    }
    \\
    \subfigure[\bfseries GAC]{
        \label{fig:cadis-gac}
        \includegraphics[width=0.175\textwidth]{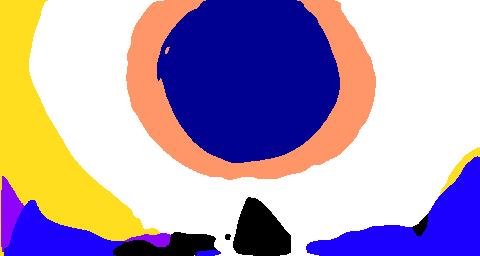}
    }
    \subfigure[SCE]{
        \label{fig:cadis-sce}
        \includegraphics[width=0.175\textwidth]{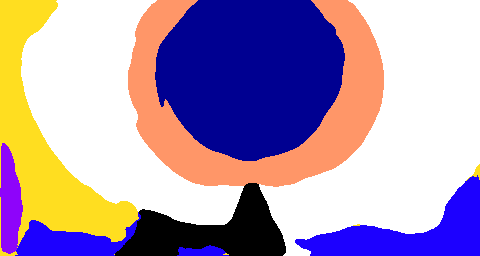}
    }
    \subfigure[\bfseries SAC]{
        \label{fig:cadis-sac}
        \includegraphics[width=0.175\textwidth]{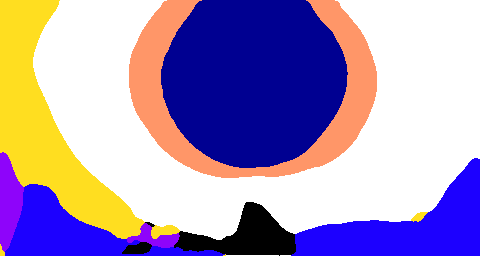}
    }
    \subfigure[Dice]{
        \label{fig:cadis-dice}
        \includegraphics[width=0.175\textwidth]{samples/cadis-dice.png}
    }
    \subfigure[\bfseries ADS]{
        \label{fig:cadis-ads}
        \includegraphics[width=0.175\textwidth]{samples/cadis-ads.png}
    }
    }
\end{figure}
\begin{figure}[htb]
    \figureconts
    {fig:dsad-vis}
    {\caption{Qualitative comparison on a challenging DSAD sample at 15\% noise. Abstaining variants (in \textbf{bold}) yield masks with better structural coherence and fewer spurious activations than their baselines.}}
    {
    \subfigure[Ground truth]{
        \label{fig:dsad-gt}
        \includegraphics[width=0.175\textwidth]{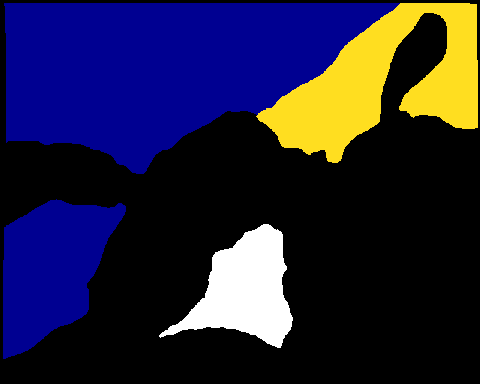}
    }
    \subfigure[CE]{
        \label{fig:dsad-ce}
        \includegraphics[width=0.175\textwidth]{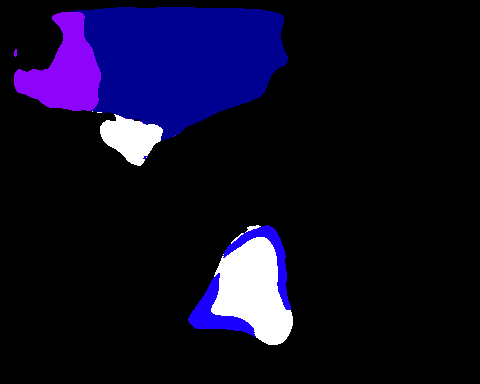}
    } 
    \subfigure[\bfseries DAC]{
        \label{fig:dsad-dac}
        \includegraphics[width=0.175\textwidth]{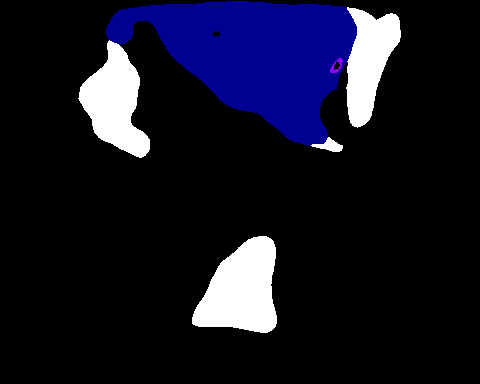}
    }
    \subfigure[\bfseries IDAC]{
        \label{fig:dsad-idac}
        \includegraphics[width=0.175\textwidth]{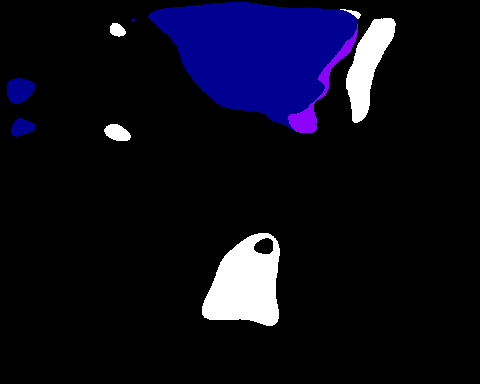}
    }
    \subfigure[GCE]{
        \label{fig:dsad-gce}
        \includegraphics[width=0.175\textwidth]{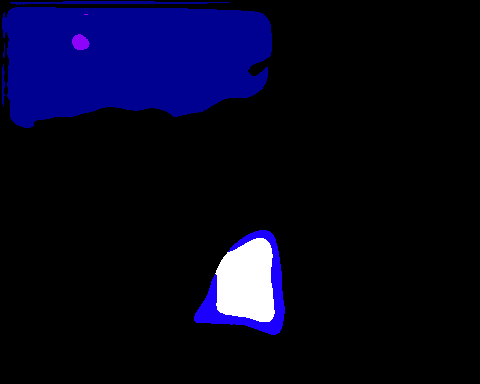}
    } 
    \\
    \subfigure[\bfseries GAC]{
        \label{fig:dsad-gac}
        \includegraphics[width=0.175\textwidth]{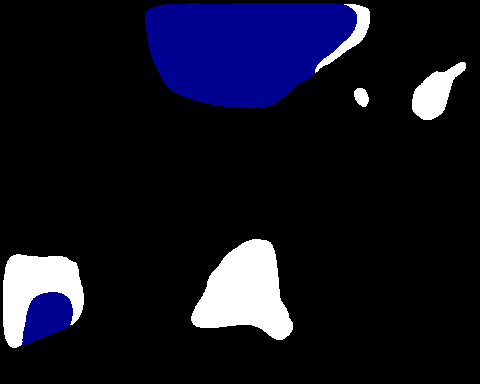}
    }
    \subfigure[SCE]{
        \label{fig:dsad-sce}
        \includegraphics[width=0.175\textwidth]{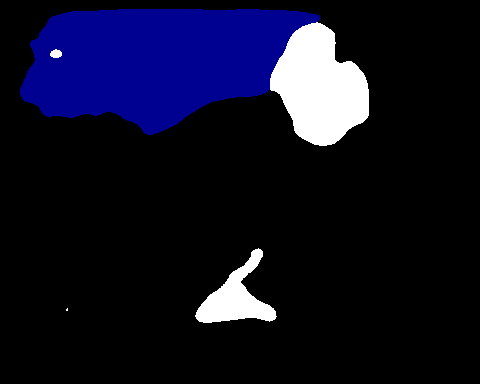}
    }
    \subfigure[\bfseries SAC]{
        \label{fig:dsad-sac}
        \includegraphics[width=0.175\textwidth]{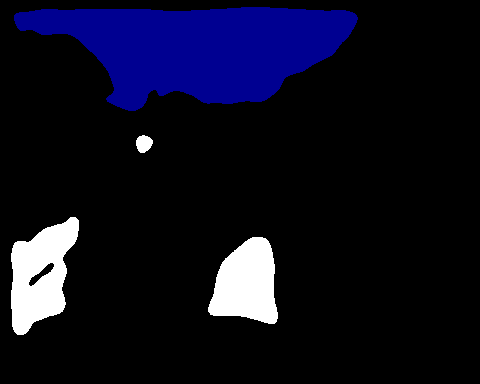}
    } 
    \subfigure[Dice]{
        \label{fig:dsad-dice}
        \includegraphics[width=0.175\textwidth]{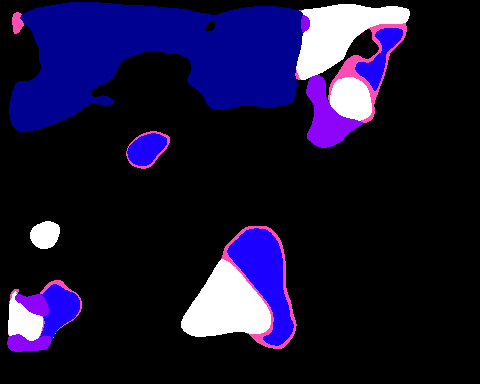}
    }
    \subfigure[\bfseries ADS]{
        \label{fig:dsad-ads}
        \includegraphics[width=0.175\textwidth]{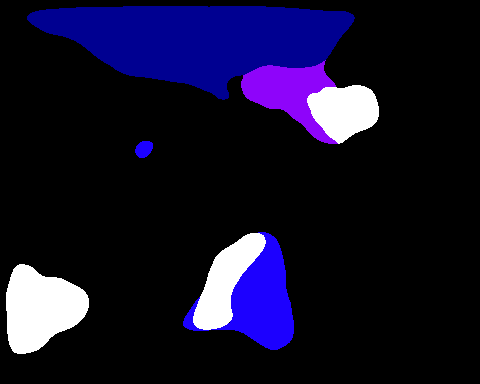}
    }
    }
\end{figure}
\pagebreak

\section{Conclusions}\label{sec:conclusions}

In this paper, we have established that the abstention mechanism is not merely a classification trick, but a fundamental and robust computer vision strategy whose capabilities are exceptionally beneficial for medical image segmentation. By introducing a universal framework equipped with informed regularization and dynamic auto-tuning, we have successfully generalized abstention to function with diverse loss paradigms, including the region-based Dice loss. This modularity allows for the creation of specialized noise-robust loss functions without requiring complex, computationally expensive changes to the underlying model architecture.

Our extensive validation on the CaDIS and DSAD datasets confirms that this approach yields consistent and significant improvements over standard baselines, particularly in high-noise situations where traditional models fail. The ability of the ADS model to produce clean, anatomically coherent masks even when trained on 25\% corrupted labels highlights the practical value of allowing models to selectively ignore unreliable supervision. 

While our results are promising, we acknowledge that our evaluation relied on synthetic noise injection, which serves as a controlled proxy for the complex, structured ambiguity often found in clinical annotations. Furthermore, the current framework relies on a pre-estimated noise rate hyperparameter. Future work will focus on validating this framework on datasets with naturally occurring inter-rater variability and developing adaptive, data-driven methods to estimate the noise rate dynamically. Ultimately, this work provides a scalable pathway toward building trustworthy diagnostic systems that can robustly learn from the imperfect data realities of the medical domain.

\clearpage 
\midlacknowledgments{This research has been funded by the Federal Ministry of Education and Research of Germany and the state of North-Rhine Westphalia as part of the Lamarr-Institute for Machine Learning and Artificial Intelligence.}
\bibliography{references}
\clearpage 
\appendix

\section{DAC \(\alpha\) auto-tuning algorithm} \label{app:alpha}
For reference and comparison, \algorithmref{alg:alpha} outlines the original linear auto-tuning schedule proposed by \citet{dac}. As illustrated, this approach requires a stateful, iterative update process where \(\alpha\) is initialized based on performance during a warm-up phase (Lines 3-9) and then incremented by a fixed \(\delta_\alpha\) at each step (Lines 16-19). In contrast, our proposed framework (\sectionref{sec:method}) simplifies this process significantly. By replacing this iterative logic with the direct power-law formulation in \equationref{eq:alpha}, we eliminate the need for state tracking and intermediate variable initialization, while simultaneously offering greater flexibility in the curriculum schedule.

\begin{algorithm2e}
\LinesNumbered
\caption{\(\alpha\) auto-tuning}
\label{alg:alpha}

\KwIn{total iter. (\(T\)), current iter. (\(t\)), total epochs (\(E\)), abstention-free epochs (\(L\)), current epoch (\(e\)), \(\alpha\) init factor (\(\rho\)), final \(\alpha\) (\(\alpha_{final}\)), mini-batch cross-entropy over true classes (\(\mathcal{H}_c(P^M_{1\dots K})\))}

\(\alpha_{set} = \text{False}\)\;
\For{\(t := 0\) \KwTo \(T\)}{
    \If{\(e < L\)}{
        \(\beta = (1-P^M_{k+1})\mathcal{H}_c(P^M_{1\dots K})\)\;
        \If{\(t = 0\)}{
            \(\tilde{\beta} = \beta\) \tcp*[r]{\{initialize moving average \}}
        }
        \(\tilde{\beta} \leftarrow (1-\mu)\tilde{\beta} + \mu\beta\)\;
    }
    
    \If{\(e = L\) \textbf{and not} \(\alpha_{set}\)}{
        \(\alpha := \tilde{\beta}/\rho\) \tcp*[r]{\{initialize \(\alpha\) at start of epoch \(L\) \}}
        \(\delta_{\alpha} := \frac{\alpha_{final}-\alpha}{E-L}\)\;
        \(update_{epoch} = L\)\;
        \(\alpha_{set} = \text{True}\)\;
    }
    
    \If{\(e > update_{epoch}\)}{
        \(\alpha \leftarrow \alpha + \delta_{\alpha}\) \tcp*[r]{\{then update \(\alpha\) once every epoch \}}
        \(update_{epoch} = e\)\;
    }
}
\end{algorithm2e}

\pagebreak
\section{Abstention Dynamics during Training}
To better understand how different loss functions utilize the abstention mechanism throughout the training process, we visualized the batch-wise abstention rate over time. \figureref{fig:abstention_dynamics} depicts the training trajectory for DAC, IDAC, and our proposed GAC on the CaDIS dataset with a synthetic noise rate of \(\eta=15\%\). The plot reveals distinct behaviours after the warm-up phase. The original DAC (blue) exhibits a rapid collapse in abstention after an initial spike. The penalty forces the abstention rate effectively to zero, meaning the model stops utilizing the mechanism and risks overfitting to noisy labels. IDAC (orange) avoids zero, but exhibits high variance. In contrast, our proposed GAC (green) demonstrates a controlled and graceful descent, eventually stabilizing at an abstention rate of approximately 15\%. 


\begin{figure}[htb]
    \floatconts
    {fig:abstention_dynamics}
    {\caption{Evolution of the abstention rate during training on the CaDIS dataset with 15\% label noise.}}
    {\includegraphics[width=0.8\textwidth]{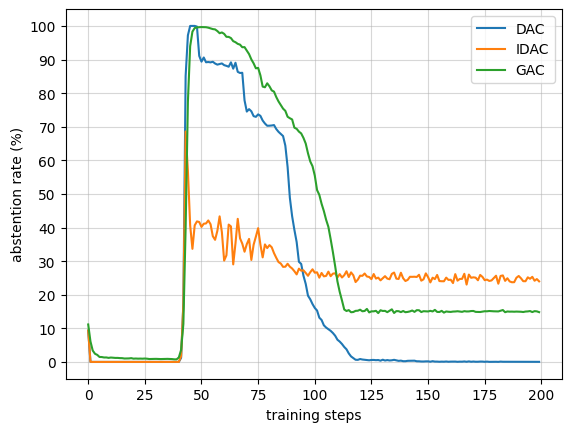}}
\end{figure}

\subsection{Alpha Auto-Tuning Behaviour}\label{app:gamma}
As described in \sectionref{sec:gamma}, our framework utilizes a power-law-based auto-tuning algorithm for the abstention penalty \(\alpha\). \figureref{fig:gamma} visually demonstrates the effect of the growth factor \(\gamma\) on the trajectory of \(\alpha\) throughout the training process, enabling sublinear, linear, or superlinear growth.
\begin{figure}[htb]
    \floatconts
    {fig:gamma}
    {\caption{The effect of different values of \(\gamma\) on the growth of \(\alpha\) with \(\alpha_{final}=1\).}}
    {\includegraphics[width=0.59\linewidth]{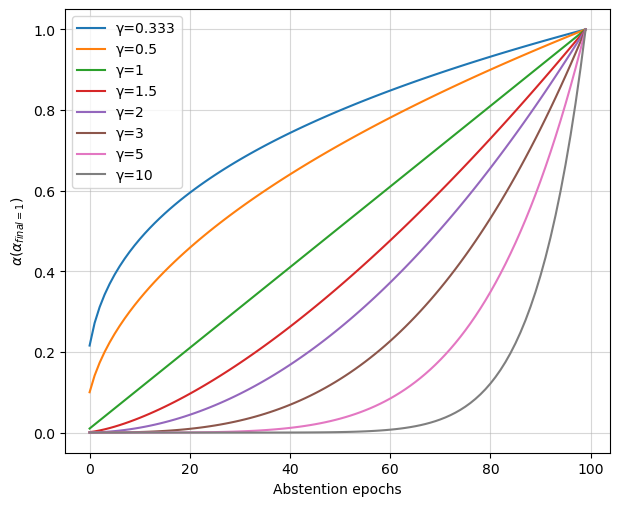}}
\end{figure}

\section{Additional Experimental Details and Method Parameters}\label{sec:appendix}
This appendix provides supplementary details regarding our experimental setup and the parameters of our proposed abstention framework.


\subsection{Global Training Hyperparameters}\label{app:train_params}
All experiments in the main paper were conducted using a consistent set of global training parameters to ensure a fair comparison. These parameters, including the network architecture, optimizer, and learning rate schedule, are detailed in \tableref{tab:train_params}.
\begin{table}[htb]
    \floatconts
    {tab:train_params}
    {\caption{Training hyperparameter configurations used in our experiments.}}
    {\begin{tabular}{@{}ll@{}}
        \toprule
        \textbf{Parameter} & \textbf{Value} \\ 
        \midrule
        Architecture & U-Net \\
        Backbone & Pretrained ResNet-50 \\
        Optimizer & AdamW \\
        Epochs & 50 \\
        Initial Learning Rate & 0.003 \\
        LR Schedule & Step decay; factor of 0.2 every 10 epochs \\
        Batch Size (CaDIS) & 128 \\
        Batch Size (DSAD) & 50 \\
        Seed Runs & 5 \\
        \bottomrule
    \end{tabular}}
\end{table}
\vfill
\subsection{Loss Function-Specific Hyperparameters}\label{app:loss_params}
The optimal hyperparameters for each loss function were determined via a sweep on the validation set at the highest noise level for each dataset. The final parameters used to generate the results in our paper are listed in \tableref{tab:loss_params}.
\begin{table}[htb]
    \floatconts
    {tab:loss_params}
    {\caption{The hyperparameter configurations for each loss function. \(L\) is the number of warm-up epochs, \(\alpha\) is IDAC's fixed abstention penalty, and \(\alpha_{final}\) is the target penalty for DAC, GAC, SAC, and ADS. \(\gamma\) is the growth factor for our enhanced \(\alpha\) auto-tuning algorithm, and \(w\) is the pooling output size for the class-wise abstention module in ADS.}}
    {
    \begin{tabulary}{\textwidth}{c|c|c|c|c|c|c|c}
        \hline
        Dataset & DAC & IDAC & GCE & GAC & SCE & SAC & ADS \\
        \hline
        CaDIS 
        & \makecell{$\alpha_{final}=1$\\$L=10$} 
        & \makecell{$\alpha=1$\\$L=10$} 
        & $q$=0.5 
        & \makecell{$\alpha_{final}=3$\\$L=10$\\$\gamma=3$} 
        & \makecell{$\alpha=1$\\$\beta=1$} 
        & \makecell{$\alpha_{final}=1$\\$L=10$\\$\gamma=1.5$} 
        & \makecell{$\alpha_{final}=1$\\$L=10$\\$\gamma=3$\\$w=16$} \\
        \hline\hline
        DSAD 
        & \makecell{$\alpha_{final}=2$\\$L=18$} 
        & \makecell{$\alpha=1$\\$L=10$} 
        & $q$=0.1 
        & \makecell{$\alpha_{final}=2$\\$L=15$\\$\gamma=2$} 
        & \makecell{$\alpha=0.5$\\$\beta=1$} 
        & \makecell{$\alpha_{final}=1$\\$L=20$\\$\gamma=3$} 
        & \makecell{$\alpha_{final}=4$\\$L=10$\\$\gamma=1.5$\\$w=16$} \\
        \hline
    \end{tabulary}
    }
\end{table}

\section{Flat}
\begin{table}[htb]
    \tableconts
    {tab:ci}
    {\caption{Quantitative analysis of robustness using the \textbf{Normalized Performance Drop Rate} ($\Delta\text{mIoU}/\Delta\eta$) accross 5 run on CaDIS and DSAD. Values represent the average mIoU points lost for every 1\% increase in label noise (Mean $\pm$ 95\% CI over 5 seeds). Lower values indicate greater resilience. 
    \textbf{Gray background:} Abstaining loss functions. 
    \textbf{(*):} Our proposed methods.
    \textbf{Structure:} The table is grouped to compare baseline losses against their abstaining counterparts.
    }}
    {
    \begin{tabular}{lcc}
        \toprule
        Loss Function & CaDIS & DSAD \\
        \midrule
        CE   & 0.950 $\pm$ 0.099 & 1.323 $\pm$ 0.379 \\
        \rowcolor{Gray} DAC  & 0.587 $\pm$ 0.167& 1.088 $\pm$ 0.346\\
        \rowcolor{Gray} IDAC & 0.687 $\pm$ 0.255 & 1.157 $\pm$ 0.149 \\
        \hline        
        GCE  & 0.711 $\pm$ 0.140 & 1.401 $\pm$ 0.166 \\
        \rowcolor{Gray} GAC* & 0.572 $\pm$ 0.140& 0.817 $\pm$ 0.197\\
        \hline
        SCE  & 0.812 $\pm$ 0.068 & 1.165 $\pm$ 0.075\\
        \rowcolor{Gray} SAC* & 0.582 $\pm$ 0.046& 1.197 $\pm$ 0.202 \\
        \hline
        Dice & 0.619 $\pm$ 0.079 & 1.108 $\pm$ 0.154 \\
        \rowcolor{Gray} ADS* & 0.426 $\pm$ 0.036& 0.803 $\pm$ 0.082\\
        \bottomrule
    \end{tabular}
    }
\end{table}

\end{document}